\documentclass{article}

\pdfoutput=1

\usepackage{arxiv}

\usepackage[utf8]{inputenc} 
\usepackage[T1]{fontenc}    
\usepackage{hyperref}       
\usepackage{url}            
\usepackage{booktabs}       
\usepackage{amsfonts}       
\usepackage{nicefrac}       
\usepackage{microtype}      
\usepackage{lipsum}

\usepackage{booktabs}
\usepackage[noend]{algpseudocode}
\usepackage{algorithm}
\usepackage{caption}
\usepackage{subcaption}
\usepackage{graphicx}
\usepackage{amsmath}
\usepackage{amssymb}

\title{Domain Adaptation for Reinforcement Learning on the Atari}

\author{
  Thomas Carr\\
  Aston University\\
  Birmingham, United Kingdom \\
  \texttt{carrtp@aston.ac.uk} \\
   \And
  Maria Chli \\
  Aston University\\
  Birmingham, United Kingdom \\
  \texttt{m.chli@aston.ac.uk} \\
  \And
  George Vogiatzis \\
  Aston University\\
  Birmingham, United Kingdom \\
  \texttt{g.vogiatzis@aston.ac.uk} \\
}

\begin{document}
\maketitle

\begin{abstract}
Deep reinforcement learning agents have recently been successful across a variety of discrete and continuous control tasks; however, they can be slow to train and require a large number of interactions with the environment to learn a suitable policy.  This is borne out by the fact that a reinforcement learning agent has no prior knowledge of the world, no pre-existing data to depend on and so must devote considerable time to exploration.  Transfer learning can alleviate some of the problems by leveraging learning done on some source task to help learning on some target task.  Our work presents an algorithm for initialising the hidden feature representation of the target task.  We propose a domain adaptation method to transfer state representations and demonstrate transfer across domains, tasks and action spaces. We utilise adversarial domain adaptation ideas combined with an adversarial autoencoder architecture.  We align our new policies' representation space with a pre-trained source policy, taking target task data generated from a random policy.  We demonstrate that this initialisation step provides significant improvement when learning a new reinforcement learning task, which highlights the wide applicability of adversarial adaptation methods; even as the task and label/action space also changes.
\end{abstract}


\section{Introduction}
  Deep Reinforcement Learning (DRL) is a successful paradigm for learning sophisticated policies through interacting with a complex environment.  DRL allows for end-to-end learning from images or raw sensor data.  RL agents generally start learning \textit{tabula rasa} with randomly parameterised neural networks and this contributes to the problem of these algorithms requiring large numbers of interactions with the environment to learn a suitable policy.  In this context, Transfer Learning is a viable route to reducing the sample complexity of these algorithms.  Many problems share similar features; an algorithm which enables an agent to identify these features quickly will, through transfer, benefit over an algorithm which has to learn these representations from scratch.

  One way to view learning within DRL is to consider the hidden layers as learning a state representation on top of which a policy can be learned.  This view has held true in the standard deep learning paradigm and the learned features are often re-purposed through direct transfer with or without fine-tuning for a subsequent task \cite{yosinski2014transferable}.  This type of approach can also work under a domain shift, however direct transfer with relatively small domain shifts can be quite detrimental to performance \cite{ganin2014unsupervised}. When the tasks to be learned are split over different input domains, the representation to be learned should be capable of describing both domains, i.e. the representation should be shared.  This representation within RL can be considered the state space.  Domain Adaptation methods in this context are about learning to construct that state space for corresponding inputs in the target domain.  This, however, raises the question of how to align observations from the two domains.

 Our method uses adversarial regularization as a mechanism to impose the source task's embedding structure on our target problem encoding space in an unsupervised manner.  If the tasks are related we expect that similar states will require similar embeddings and so we regularise our target state representation towards that of our source task.  Adversarial Auto-Encoders (AAE) \cite{makhzani2015adversarialautoencoders}, which combine an auto-encoder (AE) with a generative adversarial network (GAN) \cite{goodfellow2014generative}, provide a method to impose this regularisation in an unsupervised manner, allowing us to transfer knowledge from a source to target domain.

\begin{figure}%
\centering
\begin{subfigure}{.3\textwidth}
\includegraphics[width=0.7\linewidth,height=1.35in]{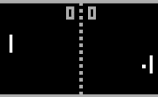}
\end{subfigure}%
\begin{subfigure}{.3\textwidth}%
\includegraphics[width=0.7\linewidth,height=1.35in]{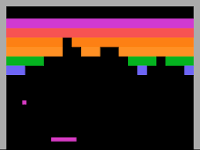}\end{subfigure}%
\begin{subfigure}{.3\textwidth}%
\includegraphics[width=0.7\linewidth,height=1.35in]{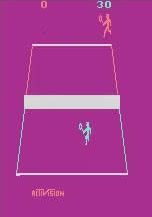}
\end{subfigure}%
\caption{\textbf{Knowledge transfer in ATARI.} Transferring skills from Pong (left) to Breakout (middle) and Tennis (right) is beyond current state of the art due to non-aligned environments and actions/rewards.}
\vspace{-0.5cm}
\end{figure}

  Our approach imposes this regularization at a distribution level and as such frees itself from the problem of aligning the data points that should share a common representation by passing this problem to the dynamics of the discriminator-generator learning problem.  Our architecture is depicted in Figure \ref{fig.Architecture} and shows how a target task encoder is trained to both reconstruct its input and lead a discriminator network into believing it was sampled from the source task encoder. We show how this approach can be successfully applied to the reinforcement learning problem across very disparate domains to provide an early performance boost over random initialization of the neural network.  

\section{Background}
  Here we present the concepts and review recent work in Reinforcement Learning, Domain Adaptation and Adversarial AutoEncoders -  which are the main building blocks of our approach.

\subsection{RL}
 RL is a machine learning paradigm centred on learning how to act in an environment.  The problem is modelled as solving a Markov Decision Process (MDP), defined by the tuple: $(S,A,T,R,\gamma)$ where $S$ is a set of states, $A$ is a set of actions, $T$ is the transition function $T:S \times A  \to \mathbb{S}$, R is the reward function $R:S \times A \times S \to \mathbb{R} $ with discount factor $\gamma$ with $0 \leq \gamma \leq 1$.  The aim is to learn a policy $\pi:S \to A$  that maximizes the discounted expected reward where the discounted expected reward is : \begin{equation} \mathbb{E}_{\pi}[\Sigma^{T}_{t=0}\gamma^tR(s_t,a_t) ].\end{equation}

In our work we make use of the A2C \cite{a2c,wang2016learning} algorithm.  A2C is a variant of the A3C \cite{mnih2016asynchronous} algorithm that performs synchronous updates.  A2C is an actor-critic algorithm so a state value function and action distribution are learned jointly; with the value function critiquing the policy.  Actor-Critic algorithms are versatile algorithms, their relationship to generative adversarial networks has been explored by \cite{pfau2016connecting} and recent work has applied them to the problem of distributed multi-task learning \cite{macua2018diffdac}.

\subsection{AutoEncoding}
  Auto-Encoders (AE) are a powerful unsupervised representation learning tool.  There are many variants, however the vanilla AE works by learning to reproduce its input, usually by passing through a lower-dimensional, bottleneck, representation. This can be done with a straightforward squared error loss function \begin{equation}L(X) = \frac{1}{N}\sum_{i=1}^{N}(x-AE(x))^{2}.\end{equation}

AEs have been used successfully in a number of RL algorithms as part of multi-step training pipelines and end-to-end paradigms.  The most obvious use of the AE is to learn a low-dimensional state representation \cite{lange2010deep,van2016stable,finn2016deep,barthlearningdeepstaterepresentation}.  Many RL algorithms are designed for low-dimensional state spaces and AEs provide a powerful unsupervised method to learn such low-dimensional representations, rather than building hand-crafted features.  On the other hand, AEs also serve as a method for initialising the weights of the RL network, in the convolutional case by learning complex filters which may be hard to learn via the reward function. 

Vanilla AE embedding spaces are not perfect, as there can be undefined regions of embedding space that lead to poor regenerations and non-smooth transitions between regeneration targets.  This problem can be alleviated through the imposition of a prior, such as a Gaussian, on the embedding space \cite{kingma2013autovariational}.

\subsection{Generative Adversarial Networks}
  Generative Adversarial Networks (GANs) are a powerful method for learning a generative model of a given domain.  They work by pitting two networks against each-other.  The Generator (G) is seeking to learn a generative model and the Discriminator (D) learns to distinguish between samples from the Generator's distribution and the True distribution.  This is shown in the loss function: 
\begin{equation}
    \begin{aligned}
        \min\limits_{G}\max\limits_{D}V(D,G) ={}\\ \displaystyle\mathop{\mathbb{E}}_{x \thicksim P_{data}(x)}[\log{D(x)}] + \displaystyle \mathop{\mathbb{E}}_{z \thicksim P_{z}(z)}[1 - \log(D(G(z)))].
    \end{aligned}
\end{equation} 
where $D(x)$ is the Discriminator's classification of real samples and $D(G(z))$ is its classification of fake samples generated by $G(z)$ where $x$ is a sample from the true distribution and $z $ is a sample from some other distribution.

Many extentions to the GAN framework have been proposed. In particular, for this work we utilize the Wasserstein GAN (WGAN) \cite{arjovsky2017wasserstein}.  The WGAN minimizes the earth-mover distance and the discriminator becomes a critic predicting continuous values for the samples rather than classifying them as real or fake.  The discriminator loss to achieve this is given by: 
\begin{equation}\begin{aligned}\min\limits_{G}\max\limits_{D}V(D,G) ={} \\
\displaystyle\mathop{\mathbb{E}}_{x \thicksim P_{data}(x)}[{D(x)}]-\displaystyle\mathop{\mathbb{E}}_{z \thicksim P_{z}(z)}[ D(G(z))].\end{aligned}\end{equation} 

\subsection{Adversarial AutoEncoders}
Combining the AE and the GAN gives us the Adversarial Autoencoder (AAE)  \cite{makhzani2015adversarialautoencoders}.  This model treats the encoder of the AE as the generator of the GAN framework; its encodings are trained to mimic the imposed prior distribution and to encode the essential domain information.  This is done by passing the encoding through both the GAN discriminator and the AE decoder.  The network is trained by the two loss functions of the AE and GAN, by interleaving steps of both optimizations.  In this way the GAN can be seen as regularizing the encoding of the AE.

\subsection{Domain Adaptation}

  Domain Adaptation is a transfer learning approach that seeks to align knowledge gained on a supervised source task with an unlabelled (or limited availability of labels) target dataset from a different domain.  These datasets usually share prediction labels so it is only the representation of the information that has changed.

Many methods seek to align the representation vector of the source and target data.  This can be done in a supervised or unsupervised manner, depending on labelling assumptions, however target data is usually limited if it does exist.  Aligning representations can be done in many ways; perhaps the simplest is to impose a constraint between the predicted representation of source and target data.

This alignment is often pursued through regularization of the embedding space as in \cite{gupta2017learning,tzeng2017adversarialdiscriminativedomain}.  These approaches show much promise, however, aligning samples from source and target must be considered.  There are various approaches to this, from using supervised data \cite{zhuang2015supervisedrepresentationlearning}, to assumed correspondences \cite{gupta2017learning}, to unsupervised approaches \cite{ammar2015unsupervised,tzeng2017adversarialdiscriminativedomain}.

\section{Architecture and Method}

\begin{figure}
\centering
\includegraphics[width=12cm,height=8.5cm,keepaspectratio]{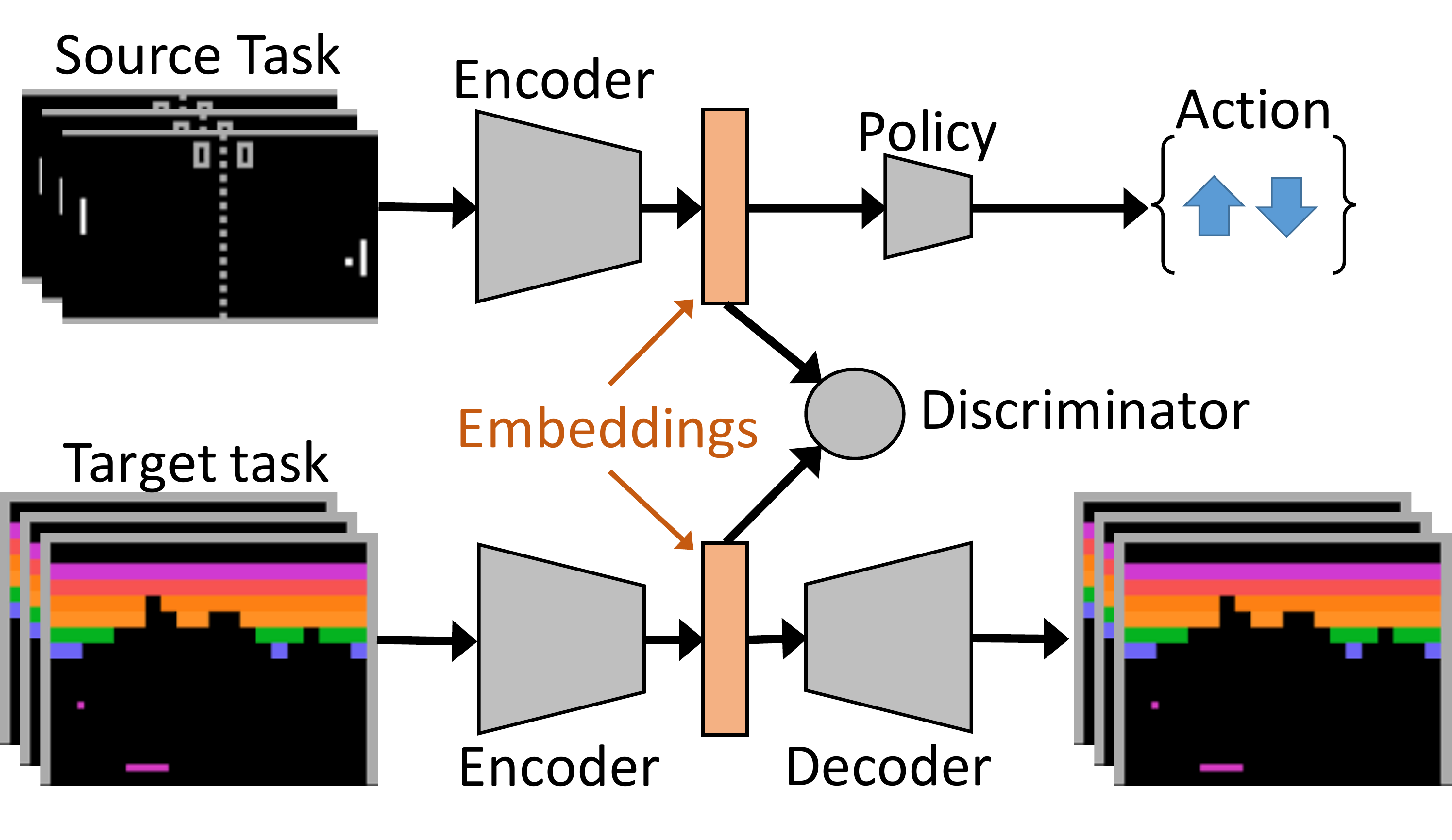}
\caption{Our agent's architecture for domain adaptation, combining adversarially discriminative domain adaptation and adversarial autoencoder architectures. A fully trained source task is depicted on top, while on the bottom is the target task autoencoder.}
\label{fig.Architecture}
\end{figure}
 Many RL methods require vast numbers of samples to learn an acceptable policy and this is caused, in part, by reliance on a reward function, which may be sparse, and by the need to explore the state space characterized by the exploration-exploitation dilemma.  Alleviating this will require many innovations; extracting as much information from the sampled data as possible is crucial.  Doing so in an unsupervised fashion will also be necessary as we do not have labels for samples gathered as the agent acts.  Another issue is that solving the problem from scratch is not a natural approach; for humans problems are often solved by drawing on knowledge already learned.  The question for RL agents then is how can an algorithm, in an unsupervised fashion, identify similarities between tasks and use that knowledge to enhance learning of subsequent problems.

Our method addresses this question by initialising the state representation of our target task to look similar to our source task. We view the hidden representation layer of our network, that is passed to the Policy layers and Value Layers, as the state representation that contains the information needed to learn how to interact with the world.  With this view we also make the observation that related game domains should require similar state features; games with a ball may for instance have a representation which contains the balls position and velocity.  Such state information is clearly something which could be shared across many input domains.  

  Domain adaptation is one way to address the problem of knowledge transfer.  In this approach we already have a model which is capable of solving a source domain and we want to adapt that knowledge so that it is applicable to a target domain.  Consider for example digit classification and the problem of solving MNIST \cite{lecun1998gradient} having already solved SVHN \cite{netzer2011reading}.  We should be able to leverage the SVHN model to solve MNIST without extra labeled data.  The Adversarial Discriminative Domain Adaptation (ADDA) \cite{tzeng2017adversarialdiscriminativedomain} approach addresses this issue by passing the new domain data through a network intialised from the source network and regularising the generated representations back towards those of the source task through the use of an adversarial domain classifier.

For SVHN to MNIST this approach seems appropriate, but as domains become more dissimilar such an obvious alignment may not be enough.  Our method will be applied to transfer between Atari games where the visual domains are vastly different when compared to the varied number domains.  Our approach therefore makes use of the Adversarial Autoencoder architecture.  This architecture allows us to learn a a feature space that captures the target domain while aligning it with the source domain.

Our proposed approach seeks to transfer knowledge in this way as a pre-training phase which aligns the representations of a trained agent with that of an initialization network that is trained to efficiently encode the target task. The alignment can be achieved in an unsupervised manner by utilising an AAE style architecture and is depicted in Figure \ref{fig.Architecture}.  The initialization network will then be used as the starting weights for training an RL agent using standard deep RL approaches.

 Our algorithm is outlined in Algorithm \ref{alg:myalogrithm} and explained in the subsequent paragraphs.

\begin{algorithm}

\caption{}\label{alg:myalogrithm}
\vspace{.3cm}
\begin{algorithmic}[1]
\Require Trained Source Policy $\pi_{\theta_{s}}(a|s)$
\Require Untrained Adversarial Auto-Encoder
\For{$i = 1 \to numEpochs$}
\For{$j = 1 \to numBatches$}
\State sample target observations: $x \sim O_t$
\State Train AutoEncoder on batch

\For{$k = 1 \to N$}
\State sample source s: $d \sim \pi_{\theta_{s}}(a_s|s_s)$
\State sample target: $x \sim \pi_{random}(a_t|s_t)$
\State Train Discriminator
\EndFor

\State sample target observations: $x \sim \pi_{random}(a_t|s_t)$
\State Train Generator on batch
\EndFor
\State

\EndFor
\State\Return Encoder/Generator

\end{algorithmic}
\end{algorithm}

To combine the ADDA approach with that of the AAE requires that we replace the regularising Gaussian distribution of the AAE with a pre-trained source embedding model.  In this work that model comes from the source policy which was trained using the A2C algorithm.  Other algorithms could be substituted here, however we leave an exploration of the best algorithm state-space for adaptation to future work.  Since the source network is fixed during training of our adaptation method and to make our embedding samples more I.I.D, we run our source agent ahead of time and generate a data-set of embeddings which we can sample from.  We also generate a data-set of target task observations using the random policy; this is again done so that we can draw samples in a more I.I.D manner.

We can now train our target embedding network.  This is done following the training procedure for an AAE.  Our training procedure has three update steps per batch. First, the AE is updated to reproduce target domain images. Second the Discriminator is updated to separate source and target domain embeddings, and third, the Generator is updated to fool the Discriminator.

Finally we use the trained target embedding network to initialise the target model and train it using A2C.  This deviates from the standard application of Adversarial Domain Adaptation methods where the source task classifier layers can be directly used on the adapted body as the source and target task remain the same and share target labels.  In this work, we do not map action spaces or policies and treat the policy and value function as functions that must be learned from scratch.

\section{Experiments}

Our experiments investigate the applicability of Adversarial Domain Adaptation to the reinforcement learning problem.
  We use the Arcade Learning Environment (ALE) \cite{bellemare13arcade} to provide a set of arcade games which are commonly used as benchmarks for Deep RL.  We interface with ALE through the OpenAi Gym  platform \cite{gym}.  Our experiments focus on transfer between pairs of games.  We begin by selecting pairs of games which we deem to be related in order to verify the effectiveness of our approach; we leave automatically identifying appropriate source tasks for future work. In this work we focus on three games: Pong, Breakout and Tennis.

  Pong is the virtualisation of Ping-Pong and is played with the ball moving sideways across the screen and the player's paddle moving along the vertical axis.  Breakout is a brick-breaking game where the player bounces a ball up to break the bricks while moving a paddle side-to-side at the bottom of the screen.  Breakout and Pong use a very similar mechanic, however, the image domain, action space, transition function and reward structure are different.

\begin{figure*}
\centering
\includegraphics[width=15cm,height=9cm]{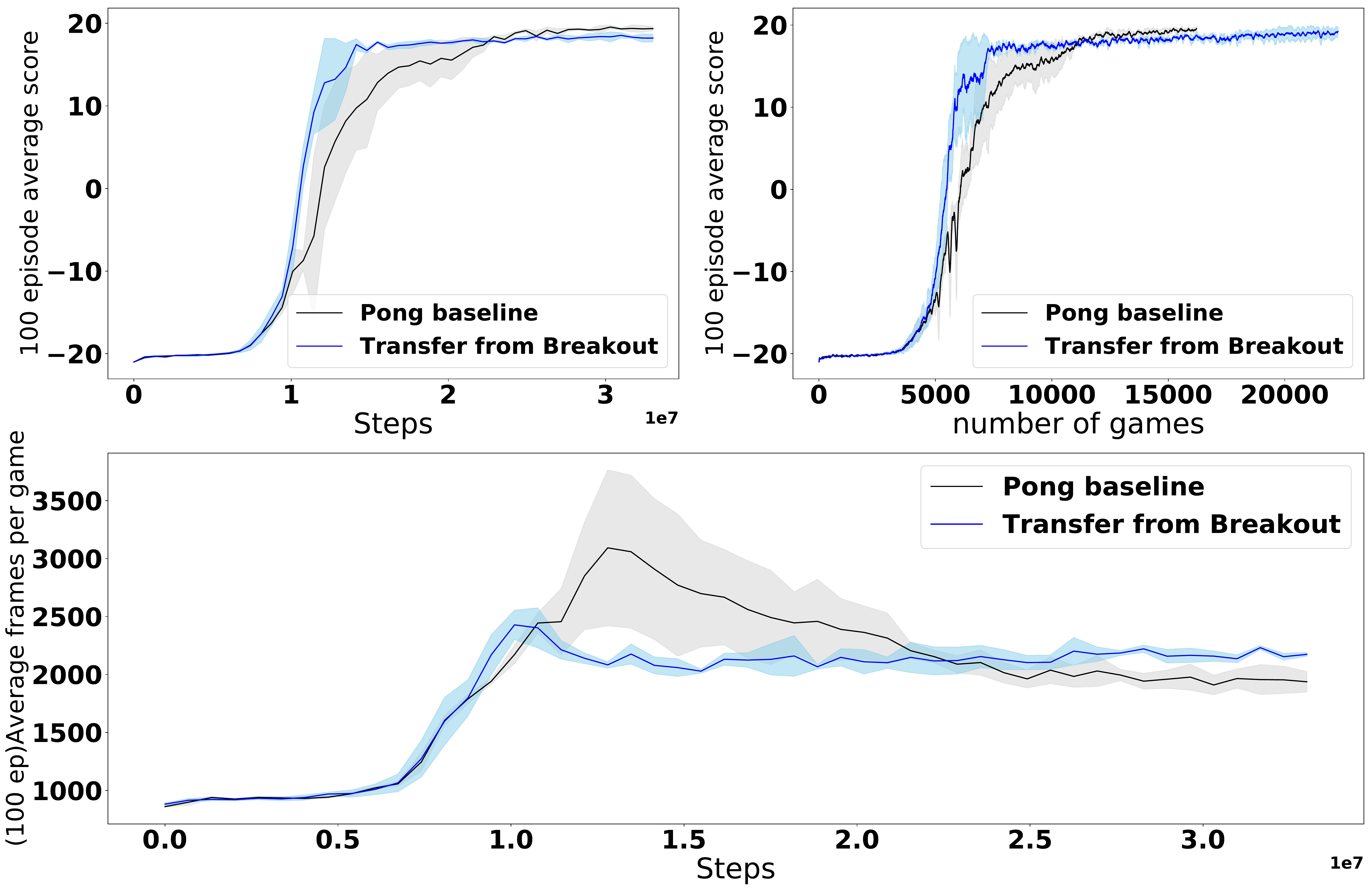}
\caption{TopLeft: Learning Curve for Pong, showing average score per game, against number of batches, each batch is 80 frames. TopRight: Plots the Average score against number of games completed. Bottom: 100 episode average number of frames in a game. Blue lines represent transfer and black represent random initialization, the filled regions represent 1 standard deviation. Source Task: Pong}
\label{pong}
\end{figure*}

  Tennis is a more difficult task than Pong or Breakout and many deep RL algorithms have reported mediocre scores \cite{mnih2016asynchronous}; although human performance reported in  \cite{mnih2015human} tends to be worse.  This game also focuses on returning a ball.  The action space for Tennis is much larger; in full it is an 18-dimensional action space, as the agent can move forward and backward to cover the whole court.  Tennis also switches the agent's sides on the court, multiple times per set, so the control of the agent becomes more complicated as the algorithm must identify which of the two agents it is playing as.  This leads to an exploration problem where the agent tends not to score well for a significant number of games.

We start with transfer from Pong (source) to Breakout (target).  We describe the experiment procedure in terms of these games, however, we follow the same procedure for the other experiments; averaging the results over five trials. 

\begin{enumerate}
\item Train agent to play Pong in the standard RL fashion; we use A2C.
\item Run the agent through a few games of Pong capturing the final hidden layer output until we have collected 100k samples. This  represents the embedding space.
\item Run a random policy through our target game, Breakout, and capture 100k frames. No learning happens at this stage.
\item Using these two data sets train the AAE.
\item Take the weights of the Generator of our AAE, to initialise the weights of a new Policy network for the target task.  The policy and value function layer weights are randomly initialised.
\item Train the policy on the target task.
\end{enumerate}

\section{Results}
\subsection{Between Pong and Breakout}

\begin{figure*}
\centering
\includegraphics[width=15cm,height=9cm]{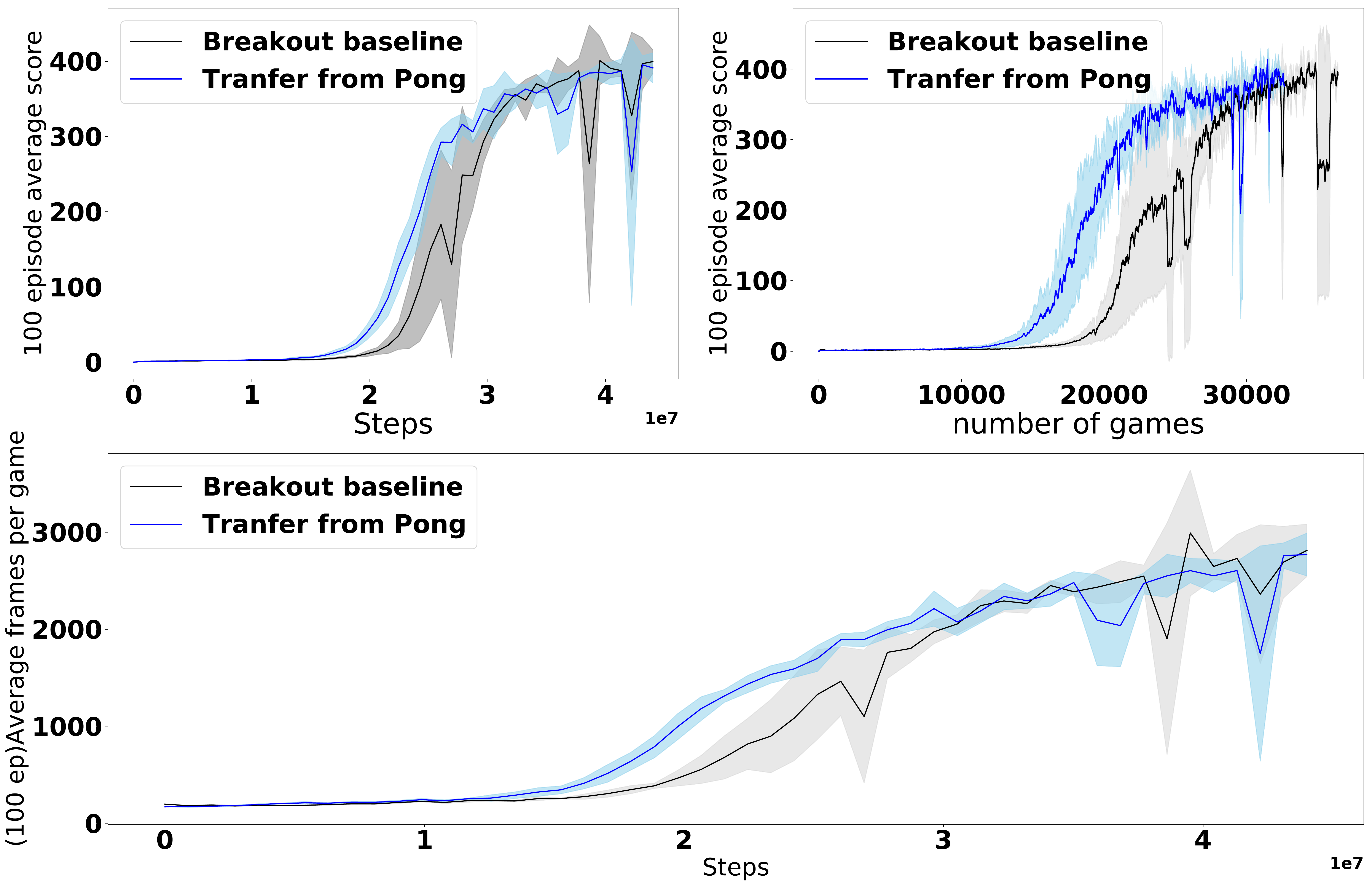}
\caption{TopLeft: Learning Curve for Breakout, showing average score per game, against number of batches, each batch is 80 frames. TopRight: Plots the Average score against number of games completed. Bottom: 100 episode average number of frames in a game.Blue lines represent transfer and black random initialization,the filled regions represent 1 standard deviation. Source Task: Breakout}
\label{breakout}
\end{figure*}

  Transferring from Breakout to Pong we can clearly see improvement from training using randomly initialised weights in Figure \ref{breakout}.  Our agent reaches a solution faster than the baseline, converging to a solution approximately 800000 samples faster.   We also note that while the adapted policy is learned faster it converges lower than the baseline.  This is an example of negative transfer and highlights how evaluating transfer is a multi-faceted problem in its own right; balancing improved learning speeds, improved initial performance and improved asymptotic performance.

Another way to look at this result is to observe how the number of frames in a game changes over time.  We show the 100 episode moving average.  What we expect is that the agent will learn to score, and thus each game will last longer as more points will be played per game. As the agent improves, the opponent will score fewer points and the number of frames will start reducing again.  For the baselines this is what we observe, with high peaks of around 3500 frames in a game.  Our transfer experiments by comparison peak a lot lower at 2500.  This observation is interesting as it suggests the agent goes straight to learning a winning policy, without passing through a sort of intermediate stage.  We can also take this into account to observe that not only does our agent need fewer games to learn, each of those games is on average shorter than the source games.  Further investigation of the policy differences around these changes could provide interesting insights;  the negative transfer observed, for instance, may be explained by the discovery of a local maximum winning policy, as despite the lower convergence our agent is still winning by 19 points which is a significant advantage.

Transferring from Pong to Breakout we can also clearly see in Figure \ref{breakout} a speed-up in the acquisition of Breakout, needing significantly fewer games (in the order of thousands) to achieve a decent score.  As the agents become better their performance evens out again and they finish at comparable levels.

Figure \ref{breakout} also shows the 100 episode moving average of game length.  We can clearly see that episode length increases faster in the transfer case, meaning that the agent is better at hitting the ball back, and that the variance is less than the baseline.
\subsection{Pong and Breakout to Tennis}

  Tennis provides a challenging domain to learn in; the opponent is a strong player and so positive rewards are sparse, even if our agent does hit the ball as the opponent is able to return it.  The large action space also creates an exploration problem.  These difficulties lead us to run our algorithm for 80 million time-steps in the Tennis environment, significantly more than the 40 million used for Pong and Breakout.
\begin{figure*}
\includegraphics[width=15cm,height=9cm]{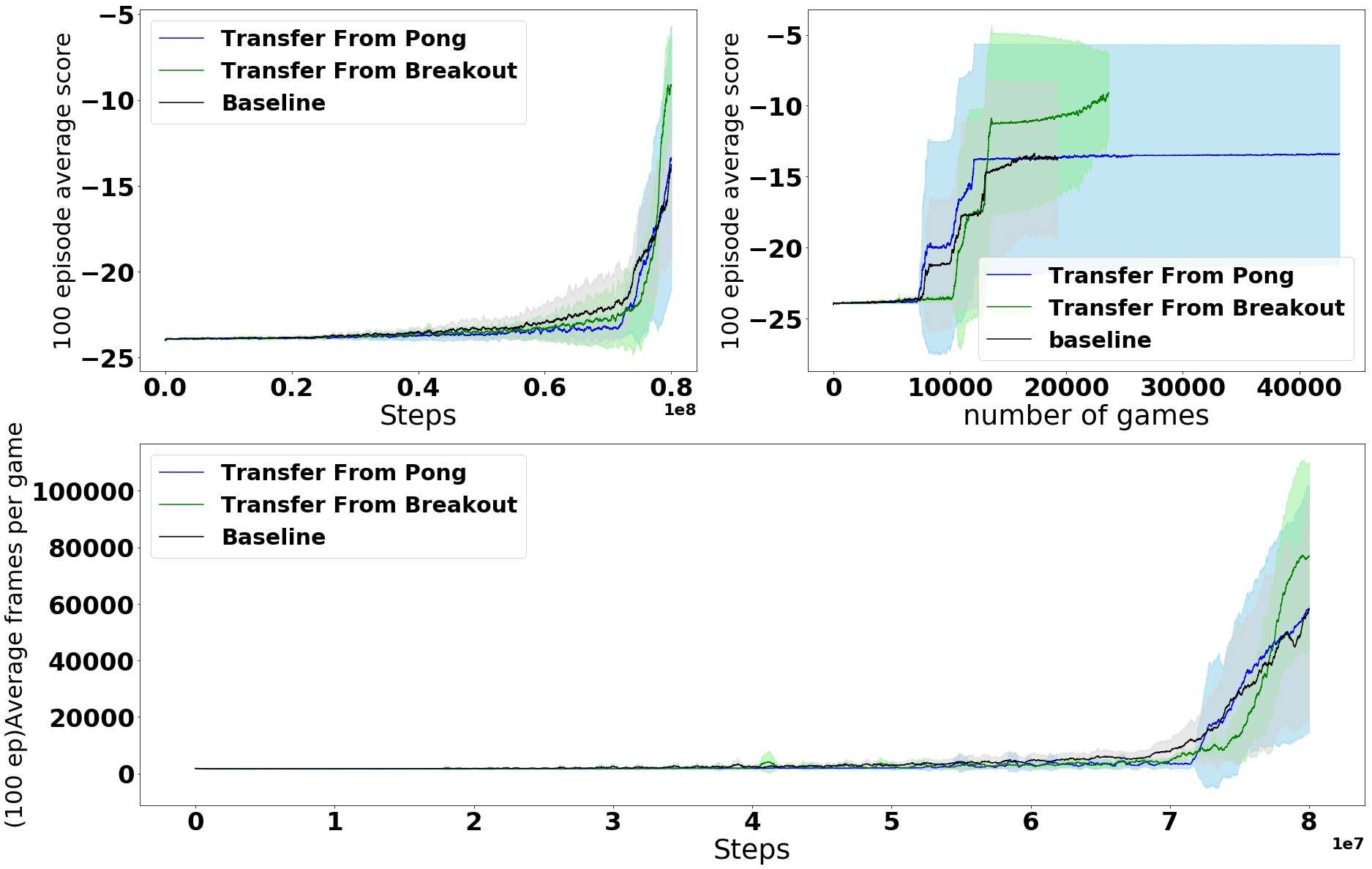}
\centering
\caption{TopLeft: Learning Curve for Tennis, showing 100 episode average score per game, against number of batches, each batch is 80 frames. TopRight: Plots the Average score against number of games completed. Bottom: 100 episode average number of frames in a game. Blue lines represent transfer from Pong, green lines transfer from Breakout and black random initialization baseline, the filled regions represent 1 standard deviation. Source Task: Breakout/Pong}
\label{Tennis}

\end{figure*}

\begin{figure}
    \centering
        \includegraphics[width=1\linewidth,height=6.5cm]{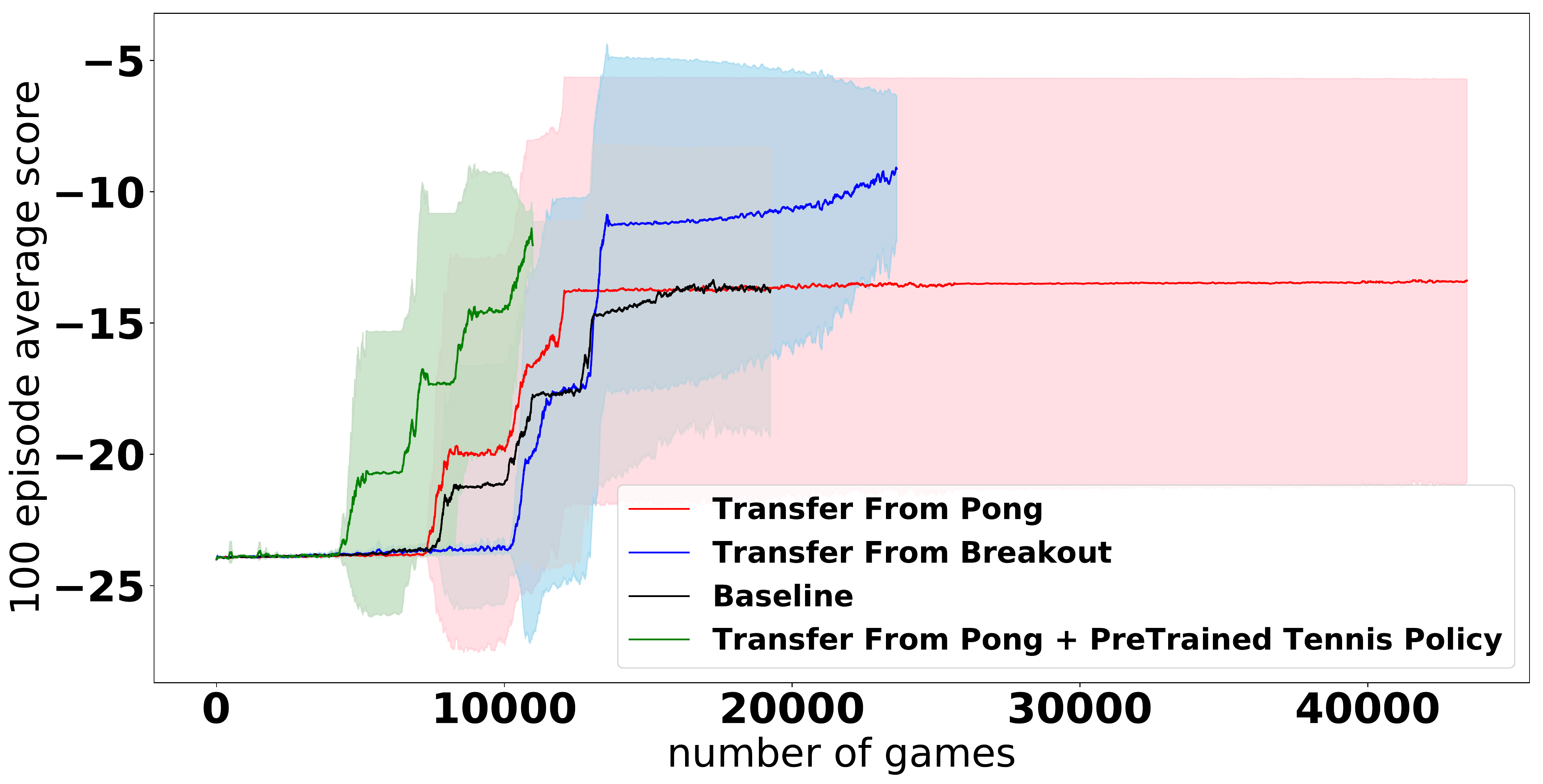}
    
\caption{Learning Curve for Tennis, showing 100 episode average score per game, against number of games. Showing transfer from Pong in red, transfer from breakout in blue and transfer from Pong, with a pre-trained policy layer and value-function layer appended.}
\label{policytrans}
\end{figure}

Figure \ref{Tennis} shows the result of transferring to Tennis from Pong or Breakout.  It is clearly difficult to learn a policy for Tennis as all our initialisations require many training steps before a scoring trend emerges.  Initialising with knowledge of Pong provides benefit by reducing the number of games the agent needs to play.

Initialising with knowledge of Breakout appears to have a different effect, slowing the agent earlier on, before taking a large learning step.  The variance of the breakout initialisations is also smaller than transfer from Pong.

\subsection{What About the Policy?}

Our proposed method focuses on learning an initial state representation utilising transferred knowledge but does not provide any policy initialisation.  Initialising the policy in conjunction with our approach could provide further improvement.  As a proof of concept demonstration of this we run a further experiment, transferring from Pong to Tennis.  In this experiment we use the same procedure as before, however, we now take the policy and value function layer from a trained Tennis model and use those to initialise the corresponding layers in the transfer model.

  Figure \ref{policytrans} shows the benefit of adding this extra policy information.  We see significant improvement, which demonstrates how our algorithm can be integrated into a larger transfer system which also predicts policy initialisations.

\section{Related Work}

 Our method is closely related to a number of AE transfer learning methods that seek to align the encoding distribution of two data-domains.  This is usually done with reference to the Kullback-Leibler (KL)-divergence between the two embedding domains which can be implicitly or explicitly minimized.  In 
\cite{zhuang2015supervisedrepresentationlearning} an AE with encoding and decoding weights shared between source and target domains is trained and the KL-divergence between their embeddings is explicitly minimised in the cost function.  Their cost function also directly incorporates source label information as a regularization term by using the embedding layer as input for a softmax prediction layer.

  In the RL context AEs can also be used as part of the initialisation process.  The approach taken in \cite{inoue2017transfer} uses Variational AEs to align real and synthetic images, for a robotics application, where the decoder is shared.  This allows them to transform one image domain into the other.  They then train an object detector on top of these reconstructed images.  

  The work in \cite{Gupta2017LearningInvariant} imposes a Kullback-Liebler divergence constraint on the encoding layer of two AEs trained in parallel.  This approach estimates an alignment through time, as two differently actuated robots solve the same task. This approach is successful but requires estimating some form of alignment between samples. 
  Our method differs from those methods by aligning encoding distributions in an unsupervised manner via Discriminator network feedback as in the AAE method.  This method utilises the Generative Adversarial Network learning algorithm to incorporate source domain information into our target domain embedding.  Our approach extends the use of AE-based transfer learning in reinforcement learning by utilising methods developed for adversarial adaptation.

  The adversarial approach we use is related to a variety of Adversarial Regularization methods.  The most related on is the AAE of \cite{makhzani2015adversarialautoencoders} where the discriminator is used as a regularisation term on the AE embedding.  We adapt this by introducing a regularisation space that transfers information from a source task.  The AAE itself is closely related to a number of approaches such as variational autoencoders (VAE) \cite{kingma2013autovariational}.  A similar approach using adversarial regularization has also been used for generating discrete data, such as text, in \cite{kim2017adversariallyregularizedae}. Adversarial approaches have also been used more explicitly for domain adaptation and \cite{tzeng2017adversarialdiscriminativedomain} provides a general framework that describes the various parts that can be changed, and what has so far been attempted.

A recent approach \cite{wulfmeier2017mutualalignment} that uses a GAN to transfer between two problems in the RL setting.  They do this by using the GAN loss for True or False samples as an extra reward variable within their RL training.  The approach explores how this reward augmentation can be used to achieve learning when the environment reward is missing or uninformative.  The domain, however, is much more aligned as they focus on transfer for robotic agents moving from simulation environments to real environments.

\subsection{Transfer and the Atari} Transfer learning  has been attempted in various forms on the Atari.  An early approach following the DQN \cite{mnih2015human} was the actor-mimic \cite{parisotto2015actor} where a multi-task atari agent was trained trying to match expert actions and representations.  This network can then be used to intialise the weights of an agent for some target task.  This pre-training mechanism initialises the filters of the target agent allowing it to start with a useful state-space.  A similar mechanic is employed in \cite{rusu2016progressive} where instead of building all relevant filters into a single network, a series of networks are joined-together feeding into the layers of each successor, adding more feature detectors.  If games share similarities then these methods allow previously learned filters to contribute to the state-space description of the new task.  The progressive neural network applied to transfer learning on the atari has a number of interesting success and failure cases and the approach comes with the down-side of continually increasing the size of the network.  A recent work (progress and compress) by \cite{schwarz2018progress} has looked at limiting the growth of progressive networks and their approach shows interesting results.

\section{Conclusion}

We have presented an adversarial method for knowledge transfer in reinforcement learning.  We have demonstrated how this approach can be used for domain adaptation to improve performance on the difficult task of learning to play Atari games.  We demonstrate how we can also change the action space and the reward function and derive benefit from the transfer approach.

Adversarial Adaptation methods offer a powerful framework for domain adaptation.  We have demonstrated how a simple application of the technique can help learning in the reinforcement learning case, even when the final layer needs to be relearned as the action space and task have changed.  This shows that the technique is not limited to simpler or more closely-related domains as previously explored such as MNIST to USPS, or SVHN to MNIST.  Recent work has sought to expand and improve the application of the technique and future work would seek to holistically integrate this technique within the RL paradigm.

As noted in related work other authors have attempted Transfer-Learning for RL and applied it in the context of the Atari.  Many of those approaches take a multi-task or lifelong learning view of the problem \cite{parisotto2015actor,kirkpatrick2017overcoming} and so have not been directly compared with.  Future work would seek to more fully explore the comparison between methods as well as understand where they fail or succeed.

Transfer for RL is a challenging domain and many recent approaches that align domains use the alignment to generate an extra reward signal for the target task agent.  Our approach demonstrates transfer across different worlds with different targets through representation alignment without any additional reward information encoded in the target task reward function.  This demonstrates the power of adversarial domain adaptation methods and provides for a variety of future research directions.

\bibliographystyle{unsrt}
\bibliography{lib}

\end{document}